\documentclass{article} 
\usepackage{nips14submit_e,times}
\usepackage{hyperref}
\usepackage{url}
\usepackage{natbib}
\usepackage{amsmath}
\usepackage{amsthm}
\usepackage{amssymb}
\usepackage{graphicx}
\usepackage{xspace}
\usepackage{tabularx}

\newcommand{\floor}[1]{\lfloor #1 \rfloor}
\newcommand{\bmx}[0]{\begin{bmatrix}}
\newcommand{\emx}[0]{\end{bmatrix}}

\newcommand{\vect}[1]{\mathbf{#1}}

\newcommand{\matr}[1]{\mathbf{#1}}

\newcommand{\vb}[0]{\vect{b}}

\newcommand{\vh}[0]{\vect{h}}

\newcommand{\vx}[0]{\vect{x}}
\newcommand{\vw}[0]{\vect{w}}

\newcommand{\vg}[0]{\vect{g}}

\newcommand{\mW}[0]{\matr{W}}

\newcommand{\mU}[0]{\matr{U}}

\newcommand{\RR}[0]{\mathbb{R}}

\newcommand{\sigmoid}{\sigma}

\nipsfinalcopy 

\title{Exponentially Increasing the Capacity-to-Computation Ratio for Conditional
Computation in Deep Learning}

\author{
KyungHyun Cho \\
Universit\'{e} de Montr\'{e}al
\And
Yoshua Bengio \\
Universit\'{e} de Montr\'{e}al \\ CIFAR Fellow
}

\begin{document}

\maketitle

\begin{abstract}
    Many state-of-the-art results obtained with deep networks are achieved with
    the largest models that could be trained, and if more computation power was
    available, we might be able to exploit much larger datasets in order to
    improve generalization ability. Whereas in learning algorithms such as
    decision trees the ratio of capacity (e.g., the number of parameters) to
    computation is very favorable (up to exponentially more parameters than
    computation), the ratio is essentially 1 for deep neural networks. Conditional
    computation has been proposed as a way to increase the capacity of a deep neural network
    without increasing the amount of computation required, by activating some
    parameters and computation ``on-demand'', on a per-example basis.  In this
    note, we propose a novel parametrization of weight matrices in neural
    networks which has the potential to increase up to exponentially the ratio
    of the number of parameters to computation.  The proposed approach is based on
    turning on some parameters (weight matrices) when specific bit patterns of
    hidden unit activations are obtained.  In order to better control for the
    overfitting that might result, we propose a parametrization that is
    tree-structured, where each node of the tree corresponds to a prefix of a
    sequence of sign bits, or gating units, associated with hidden units.
\end{abstract}

\section{Conditional Computation for Deep Nets}

Deep learning is about learning hierarchically-organized representations,
with higher levels corresponding to more abstract concepts automatically
learned from data, either in a supervised, unsupervised, semi-supervised
way, or via reinforcement learning~\citep{Deepmind-atari-arxiv2014}.
See~\citet{Bengio-Courville-Vincent-TPAMI2013} for a recent review.  There
have been a number of breakthroughs in the application of deep learning,
e.g., in speech~\citep{Hinton-et-al-2012} and computer
vision~\citep{Krizhevsky-2012-small}. Most of these involve deep neural
networks that have as much capacity (the number of units and parameters)
as possible, given the constraints on training and test time that made
these experiments reasonably feasible. 

It has recently been reported that bigger models could yield better
generalization on a number of
datasets~\citep{Coates2011-shorter,Hinton-et-al-arxiv2012,Krizhevsky-2012-small,Goodfellow+al-ICML2013-small}
provided appropriate regularization such as dropout~\citep{Hinton-et-al-arxiv2012} is used. 
These experiments however have generally been limited by training time in which
the amount of training data that could be exploited.

An important factor in these recent breakthroughs
has been the availability of GPUs 
which have allowed training deep nets at least 10 times faster, often
more~\citep{RainaICML09}.  However, whereas the task of recognizing handwritten
digits, traffic signs~\citep{Ciresan-et-al-2012} or faces~\citep{Taigman-et-al-CVPR2014} 
is solved to the point of
achieving roughly human-level performance, this is far from true for other
tasks such as general object recognition, scene understanding, speech
recognition, or natural language understanding, even with GPUs.

If we had 100 or 1000 more computing power that could be harnessed for
training, then we could train correspondingly larger models on correspondingly
larger datasets, covering more categories, modalities and concepts.  
This is important, considering that current neural network models are
still small in size (especially if we count the number of artificial
neurons) compared to biological brains, not even reaching the size of
animals such as frogs, and several orders of magnitude less than that of mammals or humans.
In this sense, we expect that much larger models are needed 
to build computers that truly master the visual world, or the
world of ideas expressed in language, i.e., to make sense of the world
around us at a level comparable to a child.

Moore's law has practically saturated if one considers only the computing power
of a single computing core. Most of the continued growth in computing power
comes from parallelization. Unfortunately, despite the impressive progress in
recent years~\citep{QuocLe-ICML2012-small,Dean-et-al-NIPS2012}, exploiting
large computer clusters to efficiently parallelize the training procedures for
deep neural networks remains a challenge. Furthermore, additionally to faster
training, in some applications we want faster inference, or test.
Thus, the question we need to ask is: besides 
distributed training, are there other ways 
to build  deep neural networks of much higher capacity 
without waiting a decade for hardware to evolve to the required level?

\citet{Bengio-tricks-chapter-2013,bengio2013estimating} have proposed
the notion of {\bf conditional computation} for deep learning to answer positively to this
question. The idea is to activate only a small fraction of the
parameters of the model for any particular examples, and correspondingly
reduce the amount of computation to be performed. 

Currently, the ratio of the number of parameters to the amount of computation is
essentially one in deep nets, i.e., every parameter is touched (usually with a
single multiply-add) for each example. In contrast, there are machine learning
models, such as decision trees~\citep{Breiman84}, with a much more favorable
ratio: with $N$ computations, a decision tree can actively select $O(N)$
parameters out of a pool of up to $O(2^N)$. Unfortunately decision trees suffer
from poor statistical properties that prevent them, like many other
non-parametric techniques relying only on the smoothness prior, from
generalizing in a non-trivial way to regions of input space far from training
examples.
    See~\citet{cucker+grigoriev99,Bengio-decision-trees10} for a mathematical
    analysis of the case of decision trees and~\citet{Bengio-2009-book} for a
    longer analysis covering a wider class of learning algorithms, such as
    Gaussian kernel SVMs and graph-based non-parametric statistical models.
On the other hand, there are both theoretical and empirical indications
suggesting that deep distributed
representations~\citep{Bengio-2009-book,Pascanu+et+al-ICLR2014b} can benefit
from advantageous statistical properties, when the data has been generated by
multiple factors organized hierarchically, with the characteristics of each
factor being learnable without requiring to see all the configurations of the
other factors.

The conditional computation for deep learning as well as this paper is
aimed at combining 
the statistical efficiency of deep learning and the computational efficiency, in
terms of ratio of capacity to computation, of algorithms such as decision trees.

With this objective in mind, we propose here a novel way to
parametrize deep neural networks (supervised or unsupervised, discriminative
or generative) that allows up to exponential increase in the ratio of 
number of parameters to computation. In other words, we allow exponentially many
parameters with respect to the amount of computation.
We achieve this by observing that 
one can exploit bit patterns associated with hidden units
in order to selectively activate different weight vectors or
weight matrices. Since the number of such bit patterns can
grow exponentially in the number of bits considered, this gives
us the required rate of growth, controllable by the maximum
size of these bit patterns.

\section{Exponentially Rich Parametrization of a Weight Matrix}

Here we consider a single layer consisting of $p$-dimensional input vector $\vx$
and $q$-dimensional output vector $\vh$. In a conventional approach, the layer
is parametrized with a weight matrix $\mW \in \RR^{p \times q}$ and a bias
vector $\vb \in \RR^q$, and computes
\begin{align*}
    \vh = \phi \left( \mW^\top \vx + \vb \right),
\end{align*}
where $\phi$ is an element-wise nonlinear function. In this case, the number of
parameters of a single layer is $O(pq)$, and very often $q=O(p)$ so the number
of parameters is $O(p^2)$.

In this note, we propose another way to parametrize a layer of a neural network,
where the number of free parameters is $O(2^k p^2)$, where $k$ is a free parameter
that controls the trade-off between capacity and computation.

Similarly to the conventional approach, a single layer consists of $\vx \in
\RR^p$ and $\vh \in \RR^q$. However, now the weight matrix is not anymore
independent of the input variable $\vx$, but is parametrized using $\vx$. The
basic idea is that $k$ bits will be derived from $\vx$ from which $O(2^k)$
weight matrices will be defined and used to define the actual weight matrix
mapping $\vx$ to $\vh$.

Let us first define a binary indicator vector $\vg \in \RR^k$ as a function
of the input $\vx$: $\vg = g(\vx)$. The gating function $g$ may
be chosen freely as long as it provides \textit{hard decisions}.
One possibility is
\[
    \vg = \left( \vx > \tau\right)_{1,\dots,k},
\]
where $\tau$ is a predefined scalar threshold. It is also
possible to make a stochastic decision such that each $g_i$ is
sampled from Bernoulli distribution with its mean
$\sigmoid(\mU^\top \vx)$, where $\mU \in \RR^{p \times k}$.

Using the binary indicators we obtain each column of the weight
matrix $\vw_j$ ($j=1,\dots,p$) as a function of $\vg$ and $j$,
using up to $k$ bits of $\vg$ (possibly chosen according to $j$) to
obtain $\vw_j$:
\begin{align*}
    \vw_j = F_j(S_j(\vg)),
\end{align*}
where $S_j$ is a subset of up to $k$ elements of $\vg$, and $F_j$
maps this binary $k$-dimensional vector to an $\RR^q$ vector of
output weights for unit $j$. For example, $F_j$ may simply bit a
look-up in a table indexed by $j$ and $S_j(\vg)$, and $S_j(\vg)$ may
simply be the first $k$ bits of $\vg$, or the set of $k$
consecutive bits of $\vg$ indexed from $\floor{j/k}$ to
$\floor{j/k}+k-1$.

One can view the above as a generalization of three-way
connections found in some
models~\citep{Memisevic+Hinton-2010,Sutskever-et-al-ICML2011} to
$k+2$-way interactions (between the $k$ gating bits, the input
$\vx_j$ and each output $\vh_i$).  For example,
\citet{Sutskever-et-al-ICML2011} select a different recurrent
weight matrix $\mW_{s_t}$ in a recurrent neural network to go
from the current state $\vh_t$ to the next state $\vh_{t+1}$
depending on the (integer) input $s_t$.

The parametrization proposed here enables the association of up
to $O(2^k)$ weight vectors with unit $j$, triggered by the
particular values of the selected $k$ bits of $\vg$. The number
of parameters is therefore $O(2^k p q)$. The required computation
depends on $F_j$, but can be as low as the cost of a table
look-up, followed by the actual computation for the matrix
multiplication, i.e., $O(pq)$.
In the next section, we describe one particular strategy of
implementing $F_j$ that aims to improve the generalization.

\section{Regularized Tree-Structured Prefix Sum of Weights}

One potential issue with the proposed scheme is that a model may
easily overfit to training samples because only a fraction of
samples are used to activate/update each of the $2^k$ possible
weight vectors. 
Beside the obvious regularization of choosing small $k$,
we propose here an additional device 
that is inspired by the impressive success of smoothed or
interpolated n-grams and back-off models in statistical language
modeling~\citep{Katz87,Jelinek80}. 

The basic idea is to maintain a set of weight vectors that are
indexed by bit sequences of different lengths. Those vectors
associated with shorter bit sequences will be updated with more
examples, therefore not requiring much regularization.
Other weight vectors 
indexed by the longer bit sequences will see few examples and be used only
to make small corrections, as needed by the data. 

As for the regularization, we simply add the norms of the weight
vectors. 
Regularization, either L1 or L2 weight decay, will 
automatically penalize more those that are less often activated,
since only when a weight vector is activated does it receive a
gradient that may counterbalance the regularizer's pull towards
0. 

We examine here one way to achieve this, based on a binary tree
structure where each node corresponds to a prefix of the $k$ bits
$\vb=S_j(\vg)$.  We thereby define $F_j(\vb)$ as follows:
\begin{align*}
  F_j(b) = \sum_{l=0}^k T(j,\vb_{1\ldots l})
\end{align*}
where $\vb_{1\ldots l}=(b_1, \ldots b_l)$ is the prefix of the
$l$ first bits of $\vb$, and $T(j,\vb_{1\ldots l})$ is a table
look-up returning an $\RR^q$ weight vector associated with unit
$j$ and bit pattern $\vb_{1\ldots l}$. With the empty sequence,
$T(j,())$ returns the default weight matrix for unit $j$.

It can be understood more intuitively by imagining a binary tree
of depth $k+1$, where each node has a weight matrix. The
above procedure traverses the tree from its root to one of the
leaves using the bit sequence $\vb$ and sums over the $j$-th
columns of the nodes' weight matrices to get the weight vector
$F_j(\vb)$.

The computation of $F_j(\vb)$ involves $O(k q)$ additions per unit
instead of being a small constant (a single table look-up), or
$O(k p q)$ in total. This is a noticeable but at the same time
reasonable  overhead over the $O(p q)$ multiply-adds that will be
required for the actual matrix multiplication.

In this case, the number of weight vectors associated with a unit
$j$ is
\begin{align*}
    \left| \theta \right| = 1+\sum_{l=1}^k 2^l = 2^{k+1}
\end{align*}
and the total number of parameters in the layer is $p q
2^{k+1}$.  However, only $2^k$ of these are actually
independent parameters while the others serve to help
regularization.  This is in contrast to the conventional case of
$O(p q)$.


Overall the degrees of freedom to computation ratio has thus
increased by $\frac{2^k}{k}$, a rapidly growing function of $k$.

As the number of parameters is much larger in the proposed
scheme, it is more efficient to implement the weight decay
regularization such that only the selected weight vectors at each
update are regularized. However, in this case, 
we must keep track of the 
interval $\Delta t = t - t'$
since each weight vector was last updated, where $t$ and $t'$ are
the current update step and the last time the weight vector was
updated. Next time the weight
vector $\vw_j$ is chosen, we treat the weight vector 
specially to compensate for the lost $\Delta t$
steps of regularization. 

For L2 weight decay, regularization
with coefficient $\lambda$ and learning rate
$\epsilon$, this simply corresponds to pre-multiplying the weight
vector by $(1-\epsilon \lambda)^{\Delta t}$:
\[
  \vw_j \leftarrow \vw_j (1-\epsilon \lambda)^{\Delta t}.
\]
This is performed before the new update is applied to
$\vw_j$.  For L1 regularization,
this can be done by moving $\vw_j$ towards 0 by $\epsilon \lambda
\Delta t$ but not crossing 0:
\[
  \vw_j \leftarrow {\rm sign}(\vw_j) \max(0, |\vw_j| - \epsilon \lambda \Delta t)
\]

\section{Credit Assignment for Gating Decisions}

One issue raised earlier by~\citet{bengio2013estimating} is the
question of training signal 
for gating decisions, i.e., the credit assignment for the gating
decisions. What is the correct way to update parameters
associated with the gating units in order 
to improve the gating decisions? 

One interesting hypothesis is that it may be sufficient to back-prop
as usual into the network by ignoring the effect of the gating
units $\vg$ on the choice of the weight vectors $\mW$. 
Although the gating decisions themselves are not adapted toward
minimizing the training loss in this case, the weight vectors are
regardlessly updated according to the objective of training. 
In other words, as long as the gating units perform a reasonable
job of partitioning the input space, it might be good enough to
adapt the exponentially many parameters stored in the table $T$.

To test that hypothesis, it would be good to evaluate alternative
approaches 
that provide training signal into the gating units.
Here are some alternatives:
\begin{enumerate}
\item Following~\citet{bengio2013estimating}
    and~\citet{Mnih+Gregor-ICML2014}, estimate a gradient using a
    variance-reduced variant of REINFORCE, i.e., by reinforcement
    learning.

\item
    Following~\citet{bengio2013estimating},~\citet{Gregor-et-al-ICML2014}
    and~\citet{Raiko2014}, estimate a gradient using a heuristic
    that propagates the gradient on $\vg$ (obtained by back-prop
    of the loss) backwards into the pre-threshold values $\vx$.

\item In the spirit of the noisy rectifier approach by
    \citet{bengio2013estimating}, compute $F_j$ as a weighted
    sum, where the gating units' activation level modulate the selected
    weight vector's magnitude:
    \[
      F_j(b,\vx) = \sum_{l=1}^k T(j,\vb_{1\ldots l}) \left(\prod_{i=1}^l (1-\tanh(x_{\pi_i}))\right)^{1/l}
    \]
    where $\pi_i$ is the index of bit $b_i$ in the input vector
    $\vx$, and $\vx$ is assumed to be the output of a rectifier,
    i.e., non-negative. Hence, when a unit is too active, it tends to
    turn off the weight contributions that it controls (which creates
    a preference for sparse $\vx$). The outside power normalizes for
    length of the controlling bit sequence.
\end{enumerate}

\section{Conclusion}

One of the greatest challenges to expand the scope of
applicability and the performance of deep neural networks is our
ability to increase their capacity without increasing the
required computations too much. The approach proposed in this
paper has the potential to achieve up to exponential increases in
this ratio, in a controllable way.

Future work is clearly required to validate this proposal
experimentally on large enough datasets for which the increased
capacity would actually be valuable, such as speech or language
datasets with on the order of a billion examples.

\bibliography{strings,strings-shorter,ml,aigaion,myref}
\bibliographystyle{natbib}

\end{document}